\documentclass{article}

\usepackage[preprint]{neurips_2025}

\usepackage[utf8]{inputenc} 
\usepackage[T1]{fontenc}    
\usepackage{hyperref}       
\usepackage{url}            
\usepackage{booktabs}       
\usepackage{amsfonts}       
\usepackage{makecell}       
\usepackage{nicefrac}       
\usepackage{microtype}      
\usepackage{xcolor}         
\usepackage{graphicx}
\usepackage{multirow}
\usepackage{tabularx}
\usepackage{ragged2e}
\usepackage{array}
\usepackage{amssymb}
\usepackage{amsmath}
\usepackage{marvosym}
\usepackage{booktabs}   

\title{AnimeShooter: A Multi-Shot Animation Dataset for Reference-Guided Video Generation}

\author{
  Lu Qiu$^{1}$,
  Yizhuo Li$^{1}$,
  Yuying Ge$^{\dagger,2}$,
  Yixiao Ge$^{2}$,
  Ying Shan$^{2}$,
  Xihui Liu$^{\dagger,1}$ \\
  \small{$^{1}$~The University of Hong Kong \quad $^{2}$~ARC Lab, Tencent PCG} \\
  \small{$\dagger$~Corresponding authors} \\
  \small{Project Page: \href{https://qiulu66.github.io/animeshooter/}{\texttt{https://qiulu66.github.io/animeshooter/}}}
}

\begin{document}

\maketitle
\renewcommand{\thefootnote}{\fnsymbol{footnote}}
\footnotetext[1]{The work was done during the author's internship at ARC Lab, Tencent PCG.} 
\renewcommand{\thefootnote}{\arabic{footnote}} 

\begin{figure}[h!]
    \begin{center}
        \includegraphics[width=0.93\textwidth]{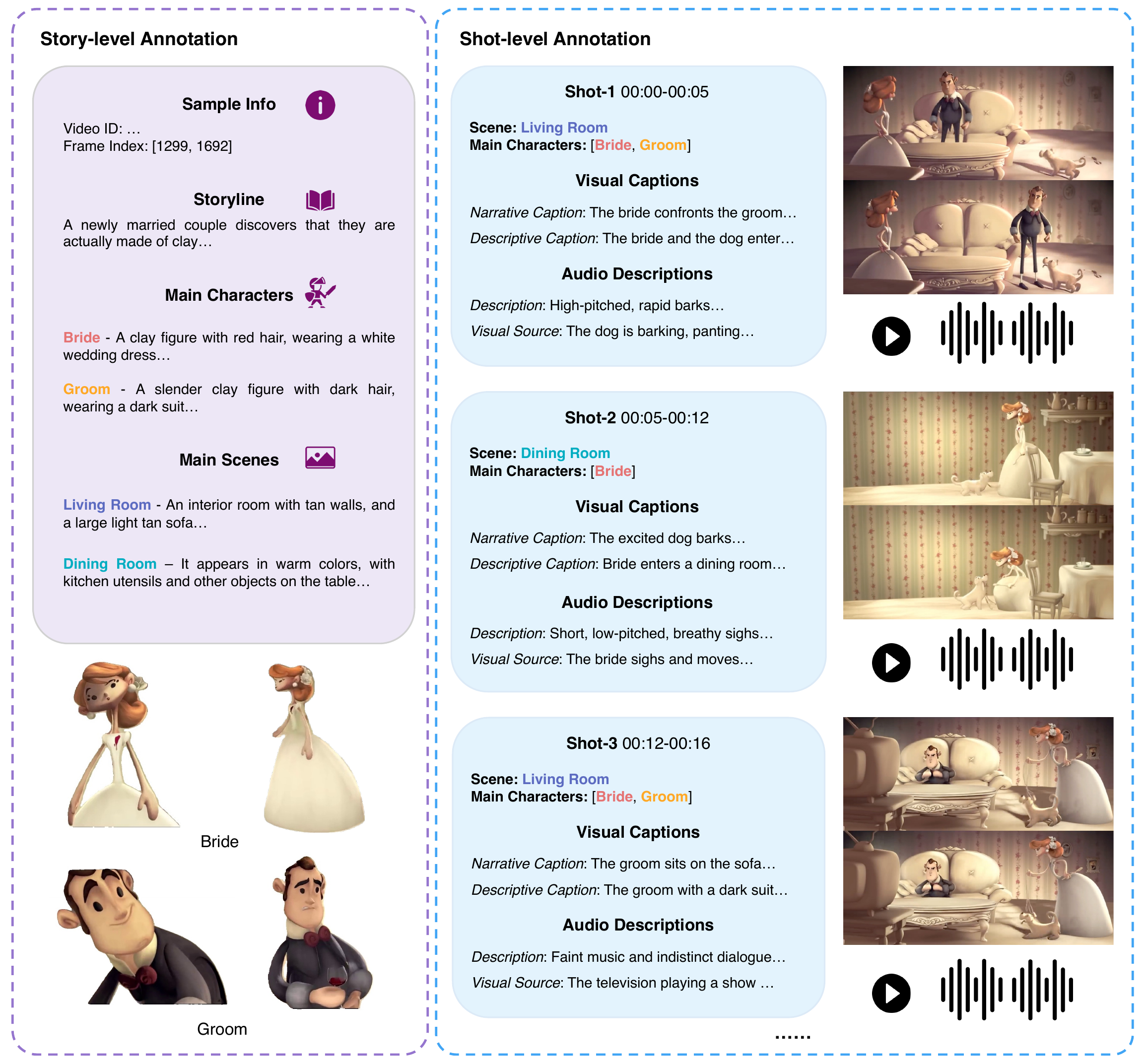}
        \caption{Overview of AnimeShooter. It is \textbf{a reference-guided multi-shot animation dataset} featuring comprehensive hierarchical annotations and strong coherence across shots. At the story level, each sample includes an overall storyline, main scene descriptions, and detailed character profiles with reference images. At the shot level, consecutive shots are annotated with specific scenes, involved characters, and rich visual captions (both narrative and descriptive). A specific subset, AnimeShooter-audio, additionally provides synchronized audios for each shot with corresponding audio descriptions and sound sources.}
        \label{fig:teaser}
    \end{center}
\end{figure}

\begin{abstract}

Recent advances in AI-generated content (AIGC) have significantly accelerated animation production. To produce engaging animations, it is essential to generate coherent multi-shot video clips with narrative scripts and character references. However, existing public datasets primarily focus on real-world scenarios with global descriptions, and lack reference images for consistent character guidance.
To bridge this gap, we present \textbf{AnimeShooter}, a reference-guided multi-shot animation dataset. AnimeShooter features comprehensive hierarchical annotations and strong visual consistency across shots through an automated pipeline. Story-level annotations provide an overview of the narrative, including the storyline, key scenes, and main character profiles with reference images, while shot-level annotations decompose the story into consecutive shots, each annotated with scene, characters, and both narrative and descriptive visual captions. Additionally, a dedicated subset, AnimeShooter-audio, offers synchronized audio tracks for each shot, along with audio descriptions and sound sources.
To demonstrate the effectiveness of AnimeShooter and establish a baseline for the reference-guided multi-shot video generation task, we introduce AnimeShooterGen, which leverages Multimodal Large Language Models (MLLMs) and video diffusion models. The reference image and previously generated shots are first processed by MLLM to produce representations aware of both reference and context, which are then used as the condition for the diffusion model to decode the subsequent shot.  
Experimental results show that the model trained on AnimeShooter achieves superior cross-shot visual consistency and adherence to reference visual guidance, which highlight the value of our dataset for coherent animated video generation.
\end{abstract}

\section{Introduction}
The animation industry plays a pivotal role in modern entertainment and education~\cite{wells2013understanding}. Recent advances in AI-generated content (AIGC) have revolutionized animation production through automated creation of complex visual narratives. Professional animation workflows necessitate the generation of coherent multi-shot video sequences that maintain visual consistency and adhere to predefined character designs. This reveals a substantial gap stemming from three fundamental limitations in existing public video datasets~\cite{bain2021frozen, yang2024vript, chen2024panda, wang2023internvid, ju2024miradata, xiong2024lvd, wang2023videofactory}: (1) focus on real-world scenario with easily obtainable web video content, (2) reliance on global captions inadequate for multi-shot narration, and (3) absence of reference images essential for consistent character guidance across sequential shots.

In this paper, we present \textbf{AnimeShooter}, a reference-guided multi-shot animation dataset featuring comprehensive hierarchical annotations and strong visual consistency across consecutive shots. Story-level annotations define an overall storyline, main scene descriptions, and detailed character profiles with reference images. The entire story is then decomposed into ordered consecutive shots. For each shot, the shot-level annotation specifies scene, involved characters, and detailed visual captions in both narrative and descriptive forms. AnimeShooter-audio is a subset which offers additional annotations of synchronized audio for each shot, along with audio descriptions and sound sources. The dataset is constructed using an automated curation pipeline as shown in Figure \ref{fig:pipeline}: we first collect and filter a diverse range of large-scale animation films sourced from YouTube, then utilize Gemini~\cite{gemini} to generate hierarchical story scripts comprising story-level and shot-level annotations. Character reference images are extracted by sampling keyframes, segmenting characters with Sa2VA~\cite{yuan2025sa2va} which is prompted by character ID/appearance, and ensuring quality with InternVL~\cite{chen2024internvl} filtering.

To demonstrate the efficacy of AnimeShooter and establish a baseline model for this challenging task, we propose AnimeShooterGen, a reference-guided multi-shot video generation model based on MLLM and diffusion model. It can generate consecutive shots in an autoregressive manner. At each generation step, both the reference image and preceding video shots are encoded by the MLLM to produce representations that simultaneously capture character identity features and visual context. We design a multi-stage training strategy to bridge the real-to-animation domain gap and achieve autoregressive multi-shot video generation. Experiments on a custom evaluation dataset comprising multiple Intellectual Properties (IPs) and extensive evaluations demonstrate that models trained on AnimeShooter effectively learn cross-shot visual consistency and adhere to specified references.

To the best of our knowledge, this is the first reference-guided multi-shot animation dataset. Through large-scale multi-shots with visual consistency, accurate reference images for character identity, and comprehensive story and shot-level annotations, we hope AnimeShooter will facilitate research and development in narrative animation generation.

\section{Related Work}
\textbf{Video-Text Datasets.} Existing text-to-video datasets present notable limitations for multi-shot animation generation. The widely adopted WebVid-10M~\cite{bain2021frozen} relies on readily available online video titles and primarily comprises short video clips. While dense-captioning datasets such as Vript~\cite{yang2024vript}, ActivityNet~\cite{caba2015activitynet}, Panda-70M~\cite{chen2024panda}, and InternVid~\cite{wang2023internvid} offer temporally segmented clips with localized descriptions, structurally akin to multi-shot annotation via timestamp-caption pairs. However, they exhibit critical narrative deficiencies, failing to maintain coherent plot progression and suffering from temporal fragmentation with abrupt gaps or redundant overlaps. In the animation domain specifically, while efforts like AnimeCeleb~\cite{kim2022animeceleb}, Sakuga-42M~\cite{pan2024sakuga}, and AniSora~\cite{jiang2024exploring} aim to build animation datasets, they are typically restricted to single-shot content, focus narrowly on character heads, or are not publicly available, thereby limiting their utility for multi-shot animation generation.

\textbf{Video Customization Datasets.} Video customization techniques facilitate the synthesis of videos centered on specific concepts, such as individuals, pets, or objects. A critical requirement for storytelling and animation generation is multi-shot video customization: the ability to synthesize a sequence of shots that maintain the consistent appearance of a predefined character. ID-Animator~\cite{he2024id} focuses on human face synthesis, utilizing facial regions as reference images. The VideoBooth dataset~\cite{jiang2024videobooth}, derived from WebVid~\cite{bain2021frozen}, augments textual prompts with image prompts generated by segmenting subjects from initial video frames via Grounded-SAM~\cite{liu2024grounding, kirillov2023segment}. Many other related efforts in video customization~\cite{chen2025multi, huang2025conceptmaster, deng2025cinema, liu2025phantom} with similar data construction pipelines are predominantly address single-shot synthesis, and target non-animation applications.

\textbf{Multi-Shot Storytelling and Animation Generation.}
Generating multi-shot videos for storytelling and animation often follows a staged pipeline. Anim-Director~\cite{li2024anim} uses image generators for reference designs, which then guide keyframe generation and subsequent I2V animation. This pipeline is shared by works like VideoStudio~\cite{long2024videostudio} and DynamiCrafter~\cite{xing2024dynamicrafter}. Except injecting reference images via multi-modal cross-attention (e.g., Anystory~\cite{he2025anystory}, VideoStudio~\cite{long2024videostudio}), some works attempt to maintain character appearance by optimization strategies (e.g., TaleCrafter~\cite{gong2023talecrafter}, DreamRunner~\cite{wang2024dreamrunner}). For example, MovieAgent~\cite{wu2025automated} leverages an LLM~\cite{grattafiori2024llama, guo2025deepseek} for script/layout generation with ROICtrl~\cite{gu2024roictrl} and ED-LoRA~\cite{gu2023mix} for character injection. Despite these advancements, the dominant per-shot generation paradigm inherently struggles with cross-shot consistency. Recent work on Long Context Tuning (LCT)~\cite{guo2025long} validates that autoregressive architectures can achieve enhanced holistic visual appearance and temporal coherence by recursively conditioning each shot on preceding visual contexts. But it also addresses real-world domains and faces limitations such as prohibitive training costs and a lack of explicit architectural mechanisms for reference image conditioning.

\section{AnimeShooter Dataset}
\begin{figure}[t]
    \begin{center}
        \includegraphics[width=1\textwidth]{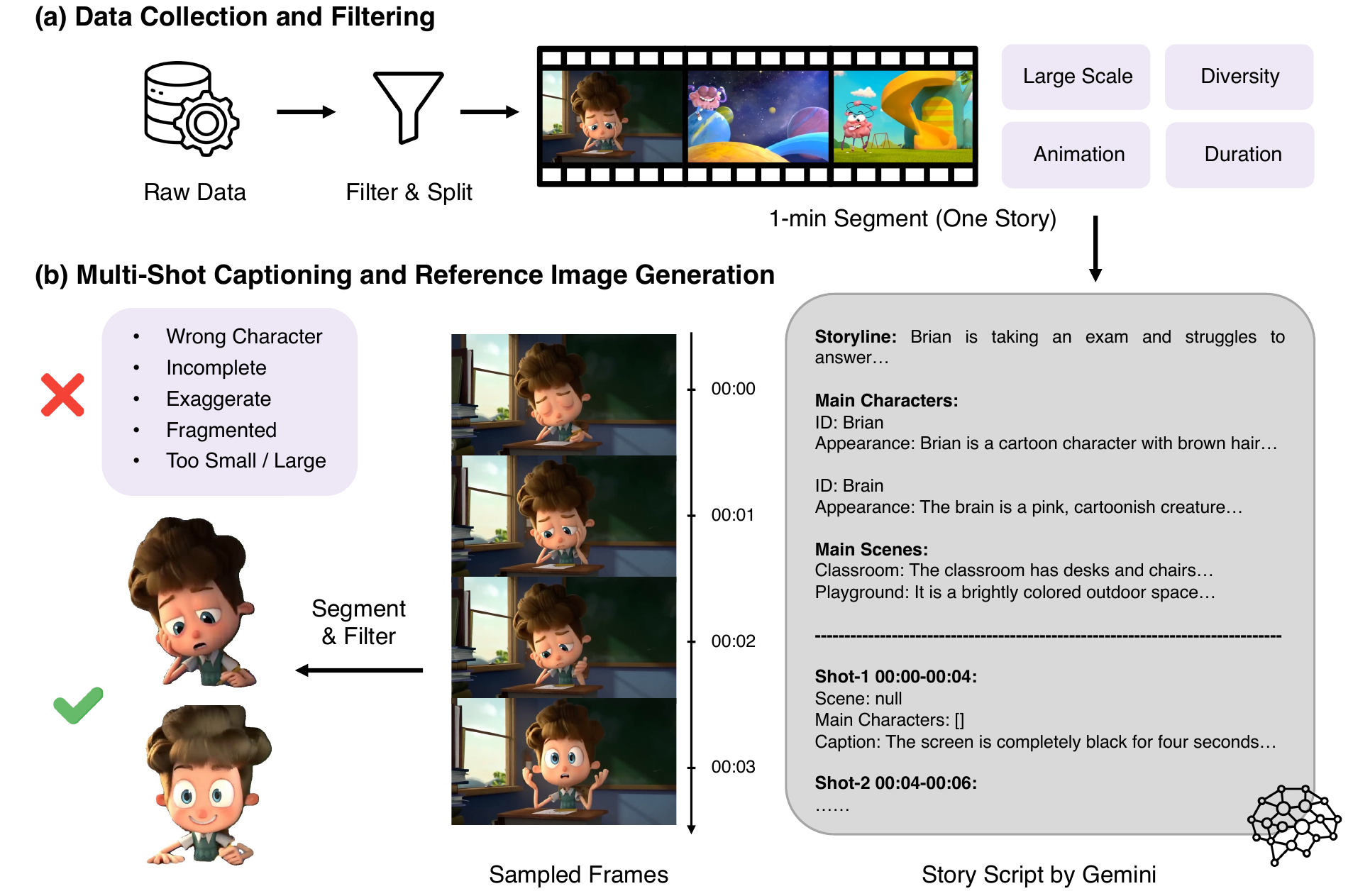}
        \caption{Video collection and annotation pipeline. We curate relevant videos from YouTube and segment them into 1-minute segments using boundary detection. Each segment serves as an individual sample representing a self-contained narrative unit (one story). We use Gemini to further decompose the story into consecutive shots with visual consistency based on transitions, and generate structured story script. Corresponding reference images are generated by Sa2VA and InternVL.}
        \label{fig:pipeline}
        \vspace{-0.5cm}
    \end{center}
\end{figure}

This section describes the generation of AnimeShooter's structured multi-shot story script and corresponding reference images. The construciton pipeline is shown in Figure \ref{fig:pipeline}. Please refer to supplementary files for the construction of additional subset AnimeShooter-audio.

\subsection{Data Collection and Filtering}
Our dataset collection begins by sourcing large-scale, diverse animated content from YouTube using keywords (e.g., "short animation", "cartoon short film"). To ensure content relevance and minimize visual-linguistic interference, we first filter these results. 16 uniformly sampled frames from each video are analyzed using InternVL~\cite{chen2024internvl} to exclude non-animated materials (such as tutorials or film reviews) and videos containing embedded subtitles. Prolonged animated content often exhibits temporal variations in character appearance, while intricate storyline with multiple characters create cognitive overload. To mitigate these issues, we implement a duration-based filtering protocol to preserve character consistency and reduce narrative complexity. Videos exceeding 20 minutes are removed firstly. The remaining videos are then cut into segments with around 1-minute durations using PySceneDetect~\cite{Pyscenedetect} algorithm to ensure coherent segment boundaries and narrative continuity. Each segment serves as an individual sample representing a self-contained narrative unit (one story).

\subsection{Multi-Shot Captioning}
\label{sec:multishot_caption}
To maintain the narrative cohesion and avoid referential ambiguity, we design a top-down multi-shot captioning strategy with three systematic phases: (1) Story-level annotation. This phase establishes a global narrative context by summarizing a succinct, coherent storyline and identifying 1-3 main characters and scenes, including detailed descriptions of their appearance and environment. (2) Shot decomposition. The entire story is subsequently decomposed into consecutive, non-overlapping shots delineated by shot transitions. (3) Shot-level annotation. For each shot, annotations identify the scene and characters, alongside two caption types: a narrative caption articulating plot progression (e.g., ``The girl said goodbye to the bear'') and a descriptive caption conveying visual details (e.g., ``A girl in red standing in front of a brown bear''). We utilize Gemini-2.0-flash~\cite{gemini} to process 1-minute segments and generate the hierarchical story script through these three phases above.

\subsection{Reference Image Generation}
\label{sec:reference}
Directly extracting frames containing a specific character is an intuitive but often inadequate strategy for obtaining reference images from animated films. The frequent co-occurrence of multiple characters and the presence of complex backgrounds can significantly hinder accurate character identification and isolation. We implements a robust model-assisted segmentation and filtering workflow. The process commences by leveraging pre-extracted story scripts to retrieve all related shots. From these shots, frames are sampled at 1 fps. Candidate frames are then fed into Sa2VA~\cite{yuan2025sa2va}, which generates initial segmentation masks based on character IDs and appearance descriptions provided as text prompts. These raw masks are refined by morphological operations to fill holes and smooth contours, contour analysis to discard masks exhibiting excessive disconnected regions, and size filtering to exclude masks that occupy less than 5\% or more than 90\% area of the image.

To guarantee the final quality of reference images, InternVL~\cite{chen2024internvl} performs a secondary verification, enforcing structural completeness of the segmented character, semantic coherence between the segmented region and the provided ID/appearance prompts, consistent appearance across instances relative to the character's textual description, avoidance of frames with extreme poses or expressions, and maintenance of high image resolution without motion blur.

\begin{table}[t]
\small
\centering
\caption{Statistics of AnimeShooter. ``Num.'' for number, ``Dur.'' for duration, ``Chars.'' for characters.}
\label{tab:statistics}
\begin{tabular*}{\textwidth}{@{\extracolsep{\fill}}cccccc} 
\toprule 
Statistics Level & Total Num. & Avg. Dur.(s) & Avg. Caption(w) & Avg. Chars. & Avg. Scenes \\
\midrule 
Video-level  & 29K & 286.57 & - & - & - \\
Story-level & 148K & 56.72 & 33.55 & 2.26 & 2.20 \\
Shot-level & 2.2M & 3.85 & 41.42 & - & - \\
\bottomrule 
\vspace{-0.5cm}
\end{tabular*}
\end{table}

\subsection{Dataset Statistics}
To ensure annotation fidelity within automated pipeline, we integrate human verification checkpoints on a small subset, validating the accuracy of story scripts and reference images. The statistical overview of the AnimeShooter dataset is presented in Table \ref{tab:statistics}. The dataset contains 29K videos, each with an average duration of 286.57 seconds. Videos are typically divided into 5.07 segments. Each segment is approximately one minute long and serves as an individual sample representing one story. These story units average 56.72 seconds and feature an average of 2.26 main characters, 2.2 main scenes, and 14.82 shots. Each shot averages 3.85 seconds and is enriched with both a 10.62-word narrative caption and a 30.8-word descriptive caption, summing to 41.42 words.

\section{Method}
\label{sec:method}
To validate the utility of AnimeShooter and establish a baseline model for animation generation, we introduce AnimeShooterGen. Inspired by prior works in world model~\cite{xiang2024pandora, huang2024owl}, AnimeShooter operates in an autoregressive fashion for reference-guided multi-shot video generation.

\subsection{Model Design of AnimeShooterGen}
\begin{figure}[t]
    \begin{center}
        \includegraphics[width=1\textwidth]{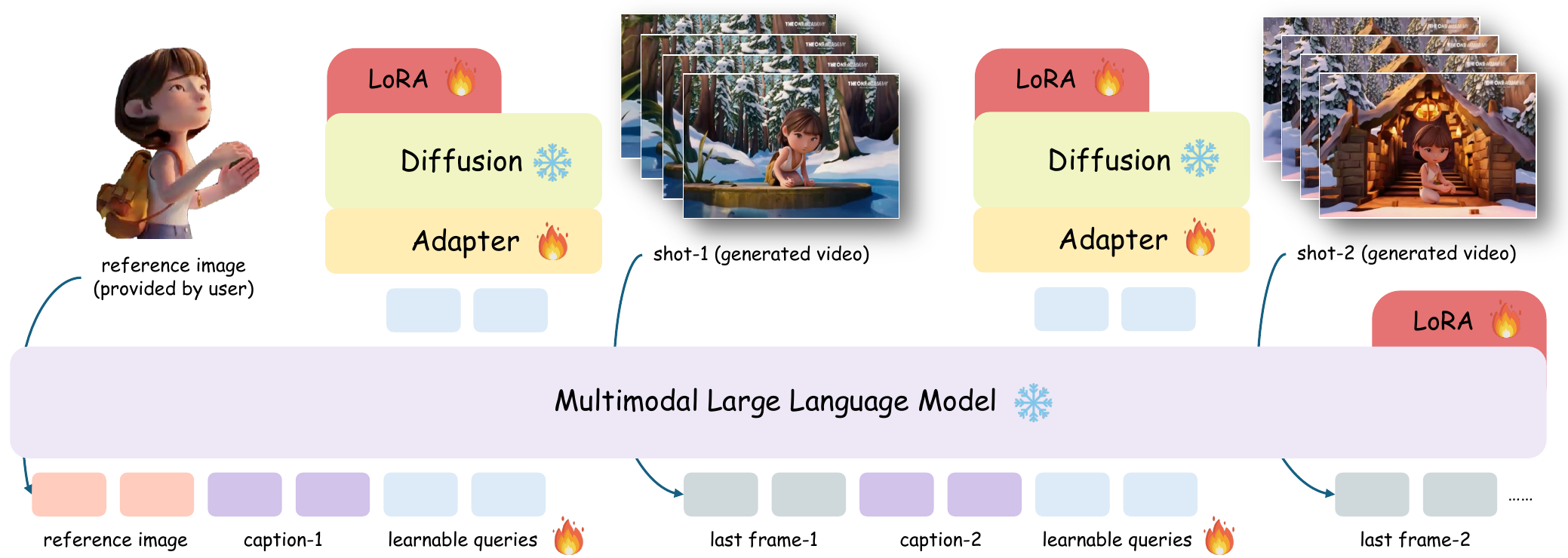}
        \caption{Overview of the model architecture. The two core components include the autoregressive backbone stemming from pretrained MLLM, and a video generator initialized from a pretrained DiT. To stitch these two components, we add a Q-Former as the adapter. This framework can generate multi-shot video in autoregressive manner.}
        \label{fig:model}
        \vspace{-0.5cm}
    \end{center}
\end{figure}

Given the character reference image $I_\mathrm{ref}$, the previous context of the story and a natural language caption for the current shot, AnimeShooterGen predicts the current $i$-th video shot, denoted as $S_i$. Figure \ref{fig:model} gives an overview of the model architecture. The model has two core components: the autoregressive backbone stemming from a pretrained MLLM~\cite{liu2024nvila} and a video generator based on pretrained DiT~\cite{peebles2023scalable, hong2022cogvideo}. An adapter (Q-Former~\cite{li2023blip}) is added to stitch these two components. For the generation of $S_i$, the MLLM backbone $f_{\text{MLLM}}$ first processes a set of inputs: a reference image $I_\mathrm{ref}$ provided by user, the accumulated previous context $C_{<i}$, and the textual caption $T_i$ for the current shot.
The previous context $C_{<i}$ encapsulates the long-term memory from preceding shots and is composed of a visual context $V_{<i}$ and a textual context $\mathcal{T}_{<i}$:
\begin{align}
    V_{<i} &= \{F_{j,\text{end}} \mid j = 1, \dots, i-1\} \\
    \mathcal{T}_{<i} &= \{T_j \mid j = 1, \dots, i-1\}
\end{align}
where $F_{j,\text{end}}$ represents the last frame of the previously generated $j$-th shot $S_j$, and $T_j$ is its corresponding caption. We set a sequence of learnable queries as the input of MLLM, and generate a conditioning signal $\text{Cond}_i$:
\begin{equation}
    \text{Cond}_i = f_{\text{MLLM}}(I_\mathrm{ref}, C_{<i}, T_i)
    \label{eq:mllm_condition}
\end{equation}
This $\text{Cond}_i$ effectively combines character visual cues from $I_\mathrm{ref}$, long-term memory from $C_{<i}$, and current textual guidance from $T_i$.
Subsequently, the video generator synthesizes the current shot $S_i$, and the training objective can be formulated as follows:
\begin{equation}
    \min_\theta \mathbb{E}_{t, x_0\sim p_{\text{data}},\epsilon\sim\mathcal{N}(0,I)} \left\|\epsilon - \epsilon_\theta(x_t,\text{Cond}_i)\right\|_2^2
\end{equation}
During the inference stage, upon the successful generation of shot $S_i$, its last frame $F_{i,\text{end}}$ and its caption $T_i$ are incorporated into the previous context to form $C_{<i+1} = (V_{<i} \cup \{F_{i,\text{end}}\}, \mathcal{T}_{<i} \cup \{T_i\})$, which is then used for generating the subsequent shot $S_{i+1}$. This autoregressive update mechanism allows the model to maintain coherence and narrative flow across multiple shots.

\subsection{Staged Training}
The training of AnimeShooterGen is conducted in a multi-stage fashion. The detailed training strategies and implementation details can be found in supplementary files.

\textbf{Condition Alignment:} The initial stage focuses on aligning the MLLM's output conditioning signal with the text encoder of the pretrained diffusion model. MLLM processes the first frame of a ground truth video clip and corresponding caption to generate an MLLM condition. We then minimize the MSE loss between this MLLM condition and the embedding of the caption extracted from the diffusion model's text encoder. In this stage, only the adapter and learnable queries are trainable.

\textbf{Single-Shot Training}: This stage aims to bridge the real-to-animation domain gap, and train MLLM to extract character visual attributes from the reference image. MLLM receives $I_\mathrm{ref}$ and $T_i$ as input, and is optimized to produce an effective condition for generating the target shot $S_{i}$. In this stage, LoRA weights of MLLM, the adapter, and learnable queries are trainable.

\textbf{Multi-Shot Training}: To foster consistency across multiple shots in terms of visual appearance, style, and color palettes, this stage extends the training to sequences. MLLM now processes the reference image $I_\mathrm{ref}$, the current caption $T_i$, and the previous context $C_{<i}$. LoRA weights of MLLM, the adapter, and learnable queries are trainable.

\textbf{LoRA Enhancement}: In all preceding stages, the core diffusion model remains frozen. To further enhance the character and style consistency provided by MLLM and refine overall video quality, this final stage involves test-time finetuning. Given a few video clips from a particular IP, we freeze all other model components and exclusively train LoRA weights added to the diffusion model.

\subsection{Integration of Audio Generation Capabilities}
\label{sec:add_audio}
To further augment the immersive quality, we integrate AnimeShooterGen with Text-to-Audio (TTA) model TangoFlux \cite{hung2024tangoflux}. The corresponding audio prompts are generated by GPT4o~\cite{gpt}. These prompts guide TangoFlux in synthesizing audios, which are then merged with the video sequences. Audio generation remains outside the scope of this paper, please refer to supplementary for details.

\section{Experiments}
\subsection{Experiment Setting}
\textbf{Baselines:} We compare two mainstream methods in storytelling field. The first approach employs a short video generation model capable of IP customization to produce individual video shots. To ensure a fair comparison and highlight the benefits of our multi-shot framework, we finetune the same pretrained diffusion model, CogVideo-2B~\cite{hong2022cogvideo}, on the same IP-specific dataset. The second approach first generates a series of IP-consistent keyframes, which are then transformed into video shots using an I2V model. Keyframe generation employs SDXL \cite{podell2023sdxl}, augmented with IP-Adapter \cite{ye2023ip} to integrate reference image features, and CogVideo-5B is utilized for the I2V conversion.

\textbf{Evaluation Dataset:} We collect 20 animation films with distinct IPs. For each IP, we manually annotate 5-6 short clips for model fine-tuning. To evaluate multi-shot generation performance, we employ DeepSeek~\cite{guo2025deepseek} to generate 10 unique narrative prompts per IP. Each prompt describes a story comprising 4 coherent shots. This process yielded a test set of 200 stories, totaling 800 video shots.

\textbf{Metrics:} Following prior works\cite{wu2025automated}\cite{cheng2025animegamer}\cite{yang2024seed}, we evaluate models using automatic metrics, advanced MLLM assessments and user studies. For automatic metrics, we employ CLIP score~\cite{radford2021learning} and DreamSim~\cite{fu2023dreamsim} to quantify the consistency between generated characters and the reference image at shot-level and story-level. We also leverage GPT-4o and Gemini 2.5 Pro as MLLM-based judges, and conduct user studies to align with human preferences. The evaluation dimensions include Overall Quality <OQ>, Character-Reference Consistency <CRC>, Multi-Shot Style Consistency <MSC> and Multi-Shot Contextual Consistency <MCC>. More detailed information are in the supplementary.

\begin{table}[t]
\small
\centering
\caption{Quantitative comparisons of automatic metrics.}
\label{tab:auto_results}
\setlength{\tabcolsep}{7pt} 
\begin{tabular}{@{}cc cccc cc@{}} 
\toprule
\multirow{2}{*}{\centering Model} & 
\multirow{2}{*}{\centering Metric} & 
\multicolumn{4}{c}{Shot-level} & 
\multicolumn{2}{c}{Story-level} \\
\cmidrule(lr){3-6} \cmidrule(l){7-8}
& & Shot-1 & Shot-2 & Shot-3 & Shot-4 & Mean & HarMeanP \\ 
\midrule
IP-Adapter + I2V & \multirow{3}{*}{CLIP $\uparrow$} & 0.8004 & 0.7814 & 0.7891 & 0.7947 & 0.7914 & 0.5901 \\
Cogvideo-LoRA & & 0.7297 & 0.7200 & 0.7417 & 0.7413 & 0.7332 & 0.5028 \\
AnimeShooterGen & & \textbf{0.8022} & \textbf{0.7949} & \textbf{0.7970} & \textbf{0.7986} & \textbf{0.7982} & \textbf{0.6121} \\[2pt]
\midrule
IP-Adapter + I2V & \multirow{3}{*}{DreamSim $\downarrow$} & 0.3679 & 0.4169 & 0.4047 & 0.3870 & 0.3941 & 0.6818 \\
Cogvideo-LoRA & & 0.4777 & 0.5060 & 0.4842 & 0.4864 & 0.4886 & 0.7759 \\
AnimeShooterGen & & \textbf{0.3484} & \textbf{0.3820} & \textbf{0.3799} & \textbf{0.3764} & \textbf{0.3717} & \textbf{0.6413} \\
\bottomrule
\end{tabular}
\end{table}

\begin{table}[t]
\small
\centering
\caption{Quantitative comparisons of MLLM evaluation and user studies.}
\label{tab:mllm_results}
\setlength{\tabcolsep}{4pt}
\begin{tabularx}{\textwidth}{@{}>{\centering\arraybackslash}p{2.5cm} *{12}{c} @{}}
\toprule
\multirow{2}{*}{Model} & 
\multicolumn{3}{c}{OQ~$\uparrow$} & 
\multicolumn{3}{c}{CRC~$\uparrow$} & 
\multicolumn{3}{c}{MSC~$\uparrow$} & 
\multicolumn{3}{c}{MCC~$\uparrow$} \\
\cmidrule(lr){2-4} \cmidrule(lr){5-7} \cmidrule(lr){8-10} \cmidrule(lr){11-13}
& GPT & Gem. & Hum. & GPT & Gem. & Hum. & GPT & Gem. & Hum. & GPT & Gem. & Hum.  \\
\midrule
IP-Adapter + I2V & 6.76 & 6.15 & 2.36 & 7.19 & 5.44 & 2.26 & 6.53 & 6.66 & 3.18 & 6.07 & 5.51 & 2.71 \\
Cogvideo-LoRA & 6.96 & 4.82 & 2.83 & 6.82 & 2.57 & 2.31 & 6.64 & 4.50 & 2.75 & 6.30 & 3.64 & 2.39 \\
AnimeShooterGen & \textbf{7.19} & \textbf{6.88} & \textbf{4.23} & \textbf{7.87} & \textbf{6.54} & \textbf{4.72} & \textbf{7.15} & \textbf{8.24} & \textbf{4.63} & \textbf{6.68} & \textbf{7.07} & \textbf{4.52} \\
\bottomrule
\end{tabularx}
\end{table}

\subsection{Quantitative Comparisons}
\label{sec:quantitative_results}
As shown in Table \ref{tab:auto_results}, automatic metrics evaluating character-reference alignment demonstrate that AnimeShooterGen outperforms both comparison methods. Notably, despite being trained on sequences of only 3 consecutive shots, AnimeShooterGen generalizes robustly to longer sequences during testing. Table \ref{tab:mllm_results} reveals additional advantages through MLLM evaluation and user studies. Beyond achieving superior CRC which also evaluates character-reference alignment, AnimeShooterGen exceeds comparison methods in MSC and MCC. Results underscore its dual strengths: (1) Enhanced reference image alignment. AnimeShooterGen achieves markedly higher character consistency than CogVideo-LoRA which shares the same diffusion architecture, proving that MLLM conditions effectively encode reference image features. (2) Cross-shot visual coherence. The MLLM’s memory mechanism retains historical context across shots, enabling high-level semantic alignment to guide the diffusion process in generating stylistically and contextually consistent new shots.

\begin{figure}[]
    \begin{center}
        \includegraphics[width=1\textwidth]{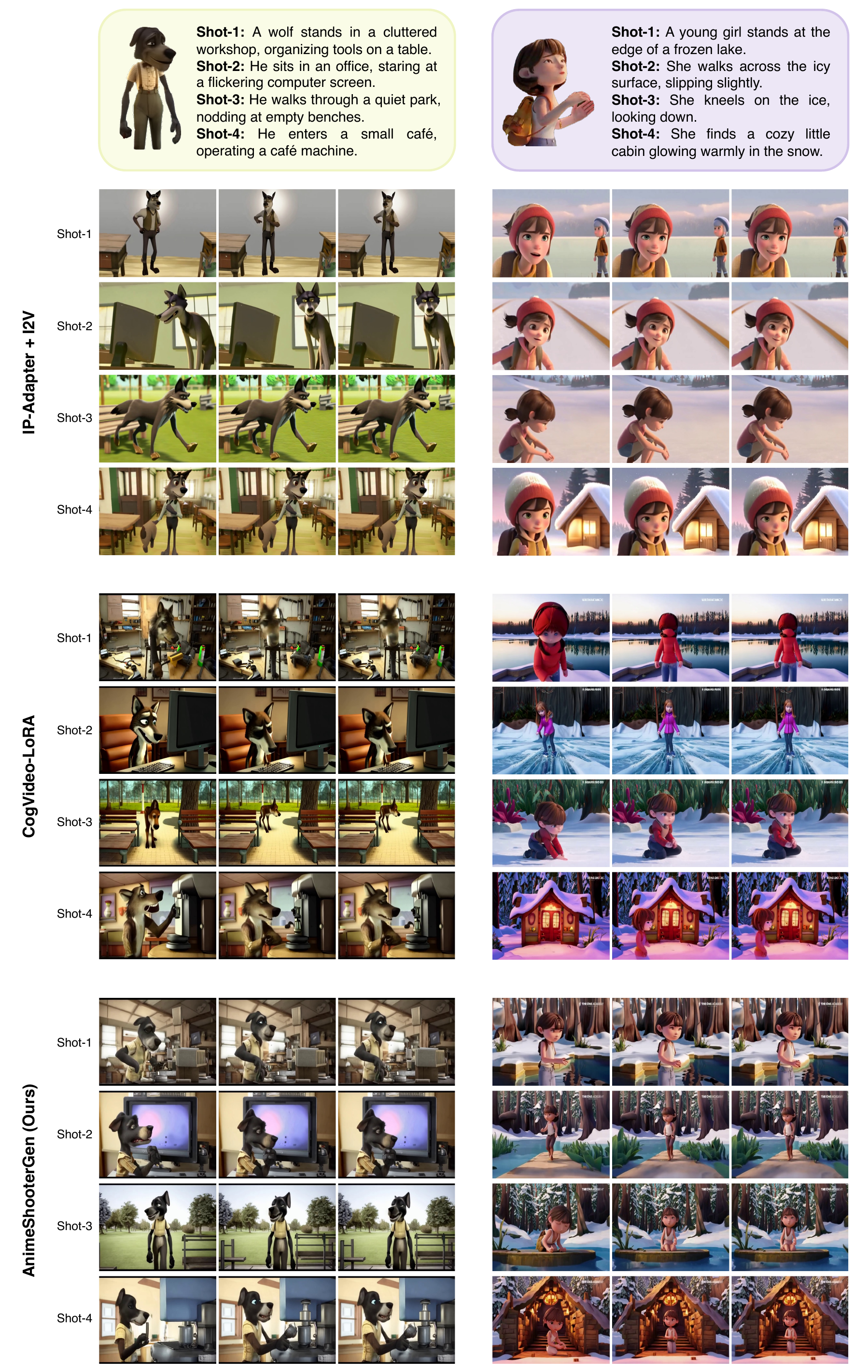}
        \caption{Qualitatively comparisons on multi-shot animation generation. Our method delivers the best visual quality, including character-reference consistency and multi-shot consistency.}
        \label{fig:experiment}
    \end{center}
\end{figure}

\begin{figure}[]
    \begin{center}
        \includegraphics[width=1\textwidth]{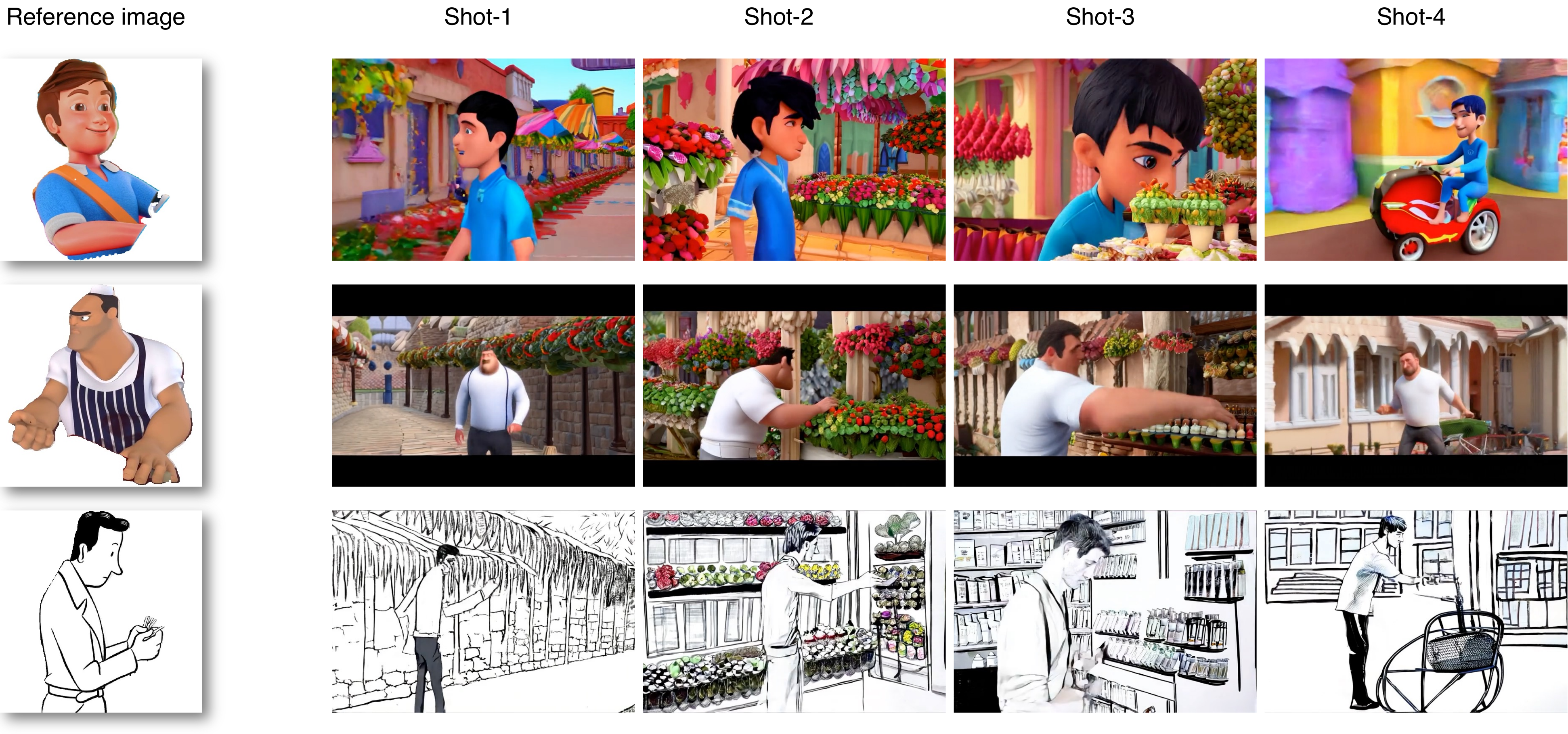}
        \caption{Visualization of using different references in MLLM \textbf{(before LoRA Enhancement)}. Shared caption: ``Shot-1: The man walks down a cobblestone street lined with blooming cherry trees, holding a vintage leather journal under his arm.'', ``Shot-2: He pauses at a flower shop, steps inside, and begins carefully selecting flowers.'', ``Shot-3: At the counter, he wraps the bouquet in paper.'', ``Shot-4: He tucks the flowers into his bicycle basket and pedals away past pastel-colored storefronts.''}
        \label{fig:reference}
        \vspace{-0.5cm}
    \end{center}
\end{figure}

\subsection{Qualitative Comparisons}
Figure \ref{fig:experiment} illustrates the visual outcomes of different methods on multi-shot animation generation task. The IP-Adapter + I2V approach struggles to maintain fidelity to the provided reference images due to weak control over IP-specific features. For instance, in the right-hand example, the generated character exhibits significant discrepancies in hairstyle, clothing, and facial structure compared to the reference image. CogVideo-LoRA also fails to achieve alignment with the reference images. Critically, both comparison methods generate individual shots as independent processes, leading to glaring inconsistencies between shots. In the left-hand example, both comparison methods depict the wolf in its natural animal form in the third shot, with anthropomorphic representations generated in the remaining shots. In contrast, AnimeShooterGen achieves superior reference fidelity, and sustains cross-shot consistency in style and environmental elements, as demonstrated by the invariant morphology of snow-covered trees across multiple shots. It contribute to the autoregressive generation strategy, where previously generated shots directly condition subsequent ones. This mechanism ensures robust style uniformity and contextual coherence throughout the multi-shot sequence.

\subsection{Investigating the Impact of Reference Images}
To isolate and understand the direct influence of reference images on the generation process, we omit the LoRA enhancement phase. The model is conditioned on same captions paired with distinct reference images. As illustrated in Figure \ref{fig:reference}, the reference images inject coarse-grained visual cues into the MLLM condition, influencing both character appearance and artistic style: in rows 1 and 2, the generated characters adopt clothing colors and silhouettes that closely correspond to their respective reference images; the minimalist sketch style in row 3 directly mirrors the reference image’s aesthetic. Crucially, even in the absence of LoRA enhancement, the autoregressive nature of our framework maintains strong multi-shot consistency. This observation underscores the inherent capability of the autoregressive architecture to enforce shot-to-shot coherence.

\section{Conclusion and Limitation}
\label{sec:conclusion}
This paper introduces AnimeShooter, a comprehensive dataset for reference-guided multi-shot animation generation, featuring comprehensive hierarchical annotations and strong visual consistency across shots. Story-level annotations provide the storyline, key scenes, and main character profiles with reference images, while shot-level annotations decompose the story into consecutive shots, each annotated with
scene, characters, and both narrative and descriptive visual captions. We also propose AnimeShooterGen which can generate reference-guided multi-shot animation in an autoregressive manner. Experiments demonstrate that being trained on AnimeShooter's multi-shot annotations promotes cross-shot consistency and adherence to predefined references. Current limitations include the restriction of AnimeShooterGen from open-domain generation due to computational demands, the requirement for test-time fine-tuning to enhance character consistency, and suboptimal audio-visual synchronization resulting from a naive zero-shot audio generation approach. We anticipate AnimeShooter will serve as a valuable resource for future work aimed at developing more robust open-domain models with improved audio-visual alignment and character fidelity.

\bibliographystyle{plain}
\bibliography{references}

\newpage
\appendix

\section{Details for Data Curation}
\subsection{Multi-Shot Captioning}
\begin{figure}[htbp]
    \begin{center}
        \includegraphics[width=1\textwidth]{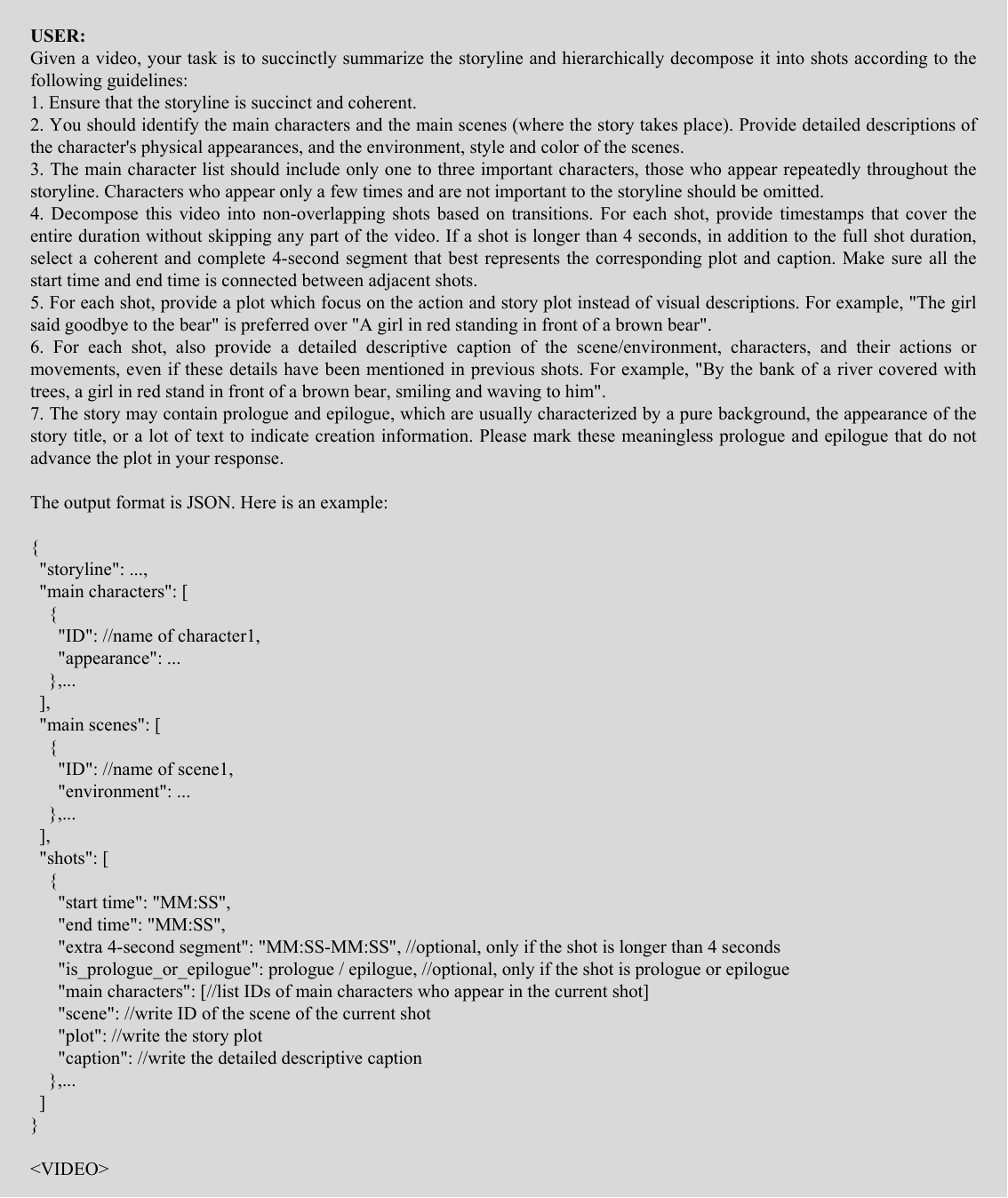}
        \caption{The prompt used for multi-shot captioning.}
        \label{fig:app_prompt_script}
        \vspace{-0.5cm}
    \end{center}
\end{figure}

In Figure \ref{fig:app_prompt_script}, we present the detailed prompt designed for multi-shot captioning (Section \ref{sec:multishot_caption}) with Gemini-2.0-flash. Initially, Gemini-2.0-flash is prompted to establish a global narrative context by summarizing a succinct, coherent storyline and identifying main characters and main scenes. Following this analysis, it decomposes the whole story into consecutive shots and provides detailed shot-level annotations.

\subsection{Reference Image Generation}
\begin{figure}[htbp]
    \begin{center}
        \includegraphics[width=1\textwidth]{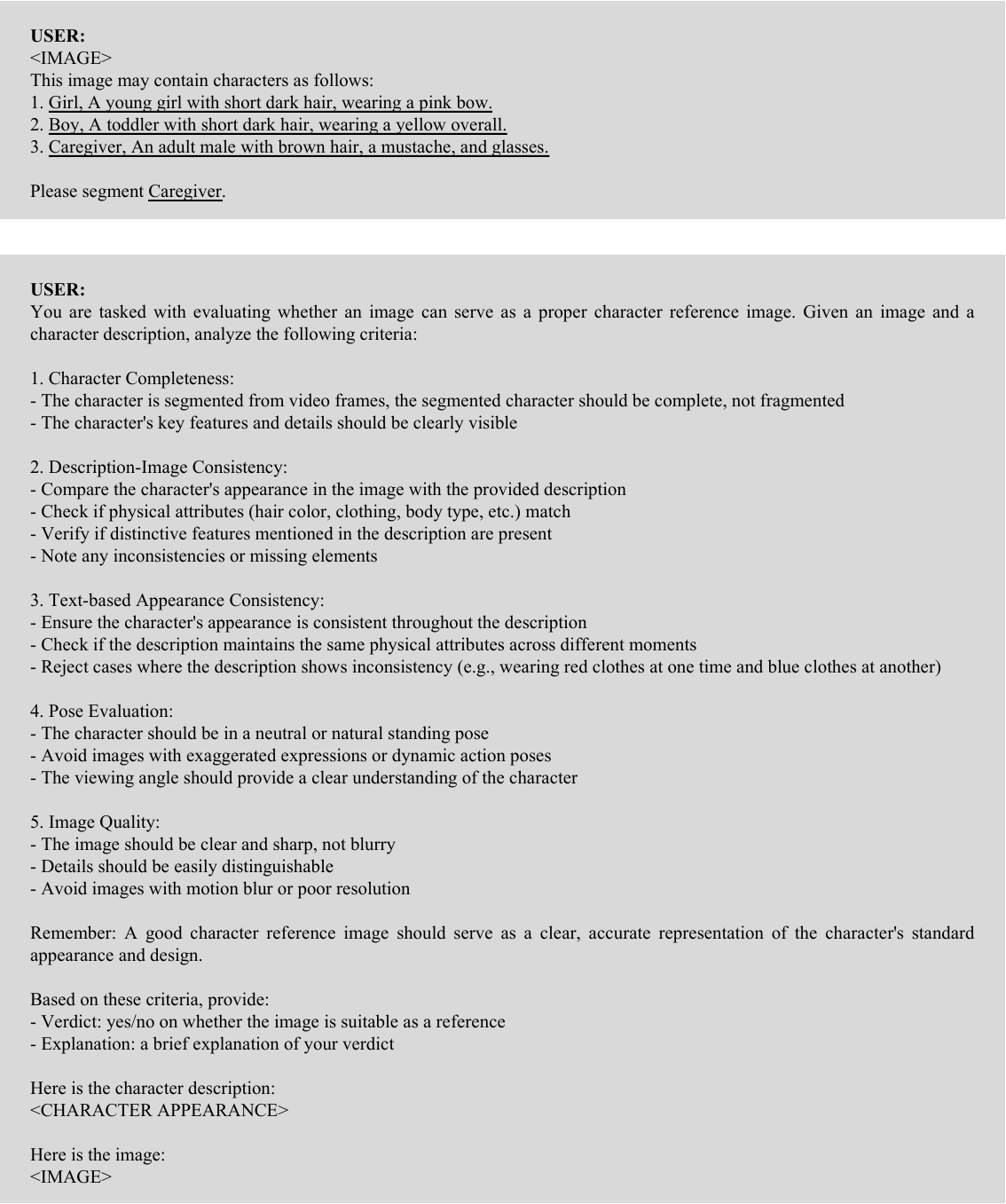}
        \caption{The prompt used for reference image generation. Top subfigure: Segmentation prompt for Sa2VA. Bottom subfigure: Filtering prompt for InternVL.}
        \label{fig:app_prompt_ref}
        \vspace{-0.5cm}
    \end{center}
\end{figure}

To address the challenge of ambiguous segmentation in images where multiple characters or non-target individuals are present, we use the prompt template for Sa2VA (Figure \ref{fig:app_prompt_ref}, top subfigure) that explicitly incorporates character IDs and appearance descriptions from the shot-level character list, followed by a clear specification of the target ID for segmentation. For post-processing, we first apply morphological opening and closing operations with a kernel size of 5 to smooth boundaries and remove noise, and then eliminate masks containing over 15 contours or 5 disconnected components. Finally, we retain only the largest connected region, and discard masks occupying less than 5\% or exceeding 90\% of the total image area. The prompt shown in the bottom subfigure of Figure \ref{fig:app_prompt_ref} is provided to InternVL for the secondary verification.

\subsection{Audio Annotation for AnimeShooter-audio}
\begin{figure}[]
    \begin{center}
        \includegraphics[width=1\textwidth]{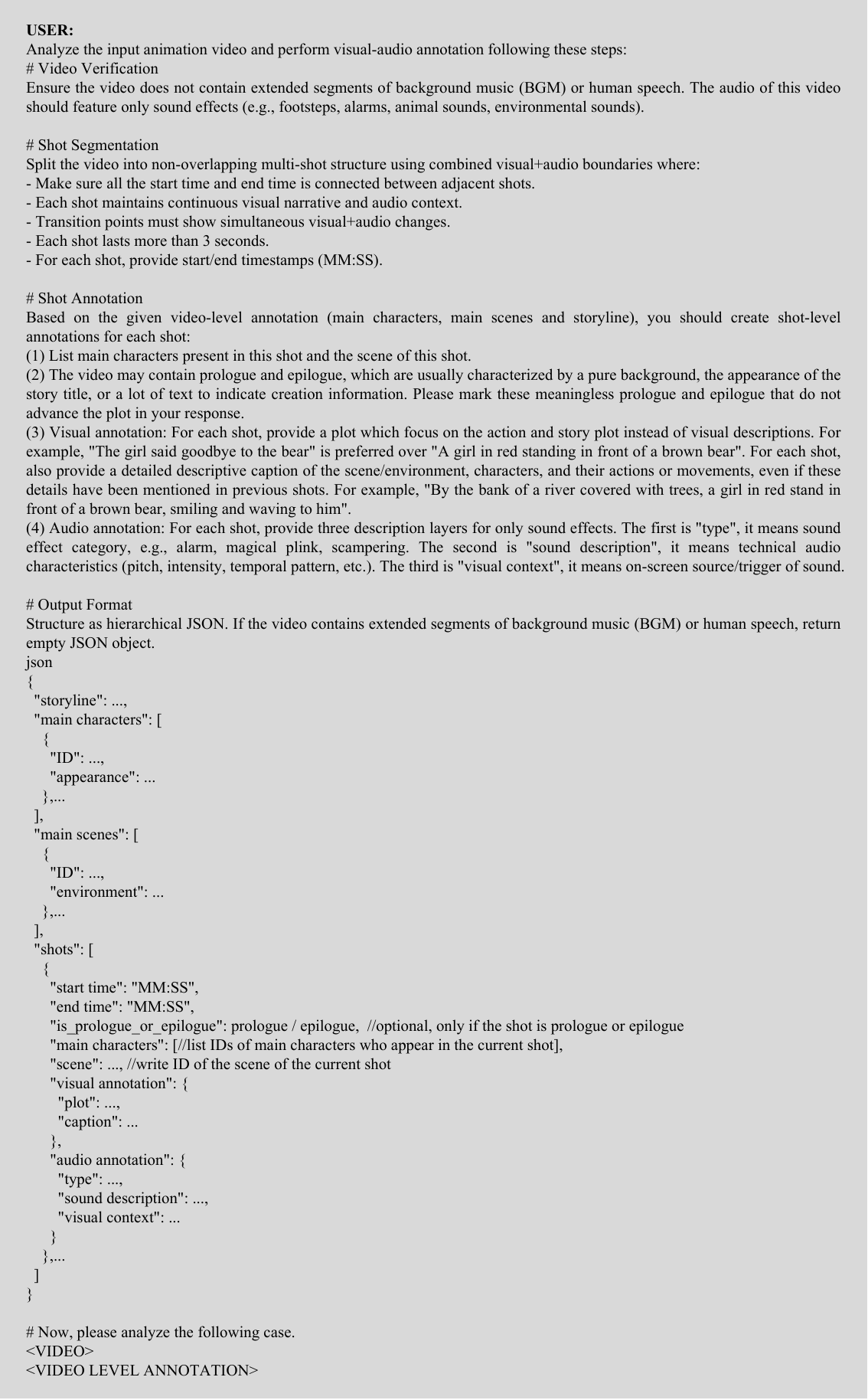}
        \caption{The prompt used for constructing AnimeShooter-audio.}
        \label{fig:app_prompt_audio}
        \vspace{-0.5cm}
    \end{center}
\end{figure}

Distinct from the visually-focused annotation in Section \ref{sec:multishot_caption}, this section aims to generate visual-audio annotations by addressing the inherent asynchrony between modalities (e.g., audio transitions lingering beyond visual shot cuts). To ground the model's analysis and maintain consistency with pre-existing holistic information, thus avoiding redundant reference image generation, video segments alongside their story-level annotations are supplied to Gemini-2.5-Pro. Gemini-2.5-Pro is required to firstly exclude video segments with prolonged background music or human speech. Following this, the model executes shot decomposition and shot-level annotation. It performs a joint analysis of visual and auditory cues to detect optimal clip boundaries where both modalities exhibit coherent transitions. For each clip, it generates: (1) two types of visual captions, (2) audio descriptors (categorizing sound types and describing tones), and (3) source attribution (mapping sounds to visual elements). Notably, these clips do not strictly adhere to single visual shot boundaries, as decomposition is determined by the joint consideration of both visual and auditory transitions. For terminological consistency, these audio-visually coherent clips are still referred to as 'shots'. The prompt used for audio annotation is shown in Figure \ref{fig:app_prompt_audio}.

\section{Details for Model Training}
\subsection{Model Framework}
AnimeShooterGen leverages NVILA-8B\cite{liu2024nvila} as the pretrained MLLM backbone and CogVideo-2B\cite{hong2022cogvideo} as the pretrained video diffusion model. An adapter, crucial for inter-model communication, is implemented using a QFormer\cite{li2023blip} with 12 layers. The length of learnable queries is set to 64. In AnimeShooterGen, for a sequence including $n$ shots, MLLM receives an input formatted as \texttt{"<Reference><Caption$_1$><LearnableQuery><Frame$_1$>...<Caption$_n$><LearnableQuery>"}, where each \texttt{<Caption$_i$>} represents textual guidance for the i-th shot and \texttt{<LearnableQuery>} serves as a placeholder for contextual feature extraction. During training, the \texttt{<Frame$_1$>} to \texttt{<Frame$_{n-1}$>} tokens are populated with the last frames from their corresponding ground truth video shots, and all $n$ shots contribute jointly to the diffusion loss through backpropagation. At inference time, the model operates autoregressively: it first generates a 1-shot sequence using only the initial reference, then replaces the \texttt{<Frame$_1$>} token with the final frame of the newly generated shot to condition the next iteration, recursively extending the sequence until reaching the target length $n$.

\subsection{Implementation Details}
For Condition Alignment, we train the model using video-caption pairs from the large-scale WebVid-10M dataset\cite{bain2021frozen}. MLLM takes the first frame of a ground-truth video clip, its corresponding caption, and learnable queries as inputs. We minimize the MSE loss between the MLLM’s output condition and the T5 features of the caption. During this phase, only the adapter and learnable queries are trainable, optimized with a batch size of 32 for 1.8 M steps. 

For Single-Shot Training, we utilize samples from the AnimeShooter dataset containing reference images and descriptive captions for individual shots. Here, the adapter, learnable queries, and LoRA parameters of the MLLM are fine-tuned with a batch size of 24 for 17K steps. Classifier-Free Guidance (CFG) is applied to enhance multimodal control, with independent dropout probabilities of 0.05 for the reference image, caption, or both modalities.

For Multi-Shot Training, we curate 3-shot sequences from the AnimeShooter dataset. The same trainable components (adapter, learnable queries, and MLLM LoRA) are updated with a batch size of 8 for 8K steps. A simplified CFG strategy is adopted, where both the reference image and caption inputs are simultaneously replaced with blank content at a 0.05 probability, eliminating modality-specific dropout. 

For LoRA enhancement, given 5-6 separate video clips from a particular IP, these are randomly sampled and combined to form training sequences of 3 consecutive shots. This targeted finetuning is performed for 1K to 2K steps with batch size of 2.

\section{Details for Experiments}
\subsection{Baselines}
For fair comparison with CogVideo-LoRA, we fine-tune the same pretrained diffusion model (CogVideo-2B) on the same IP-specific dataset and iterations as AnimeShooterGen. While AnimeShooterGen is trained in multi-shot mode (batch size=2 per step, with each sample containing 3 shots, totaling 6 shots per step), CogVideo-LoRA supports only single-shot training. To match the computational scale, we set its batch size to 6 (6 shots per step). For the training-free baseline IP-Adapter + I2V, we utilize stable-diffusion-xl-base-1.0 with its IP-Adapter to generate keyframes conditioned on reference images and captions. These keyframes and captions are then fed to the CogVideo-5B I2V model (replacing CogVideo-2B due to its lack of I2V capability) to synthesize video shots.

\subsection{Evaluation Dataset Construction}
\begin{figure}[htbp]
    \begin{center}
        \includegraphics[width=1\textwidth]{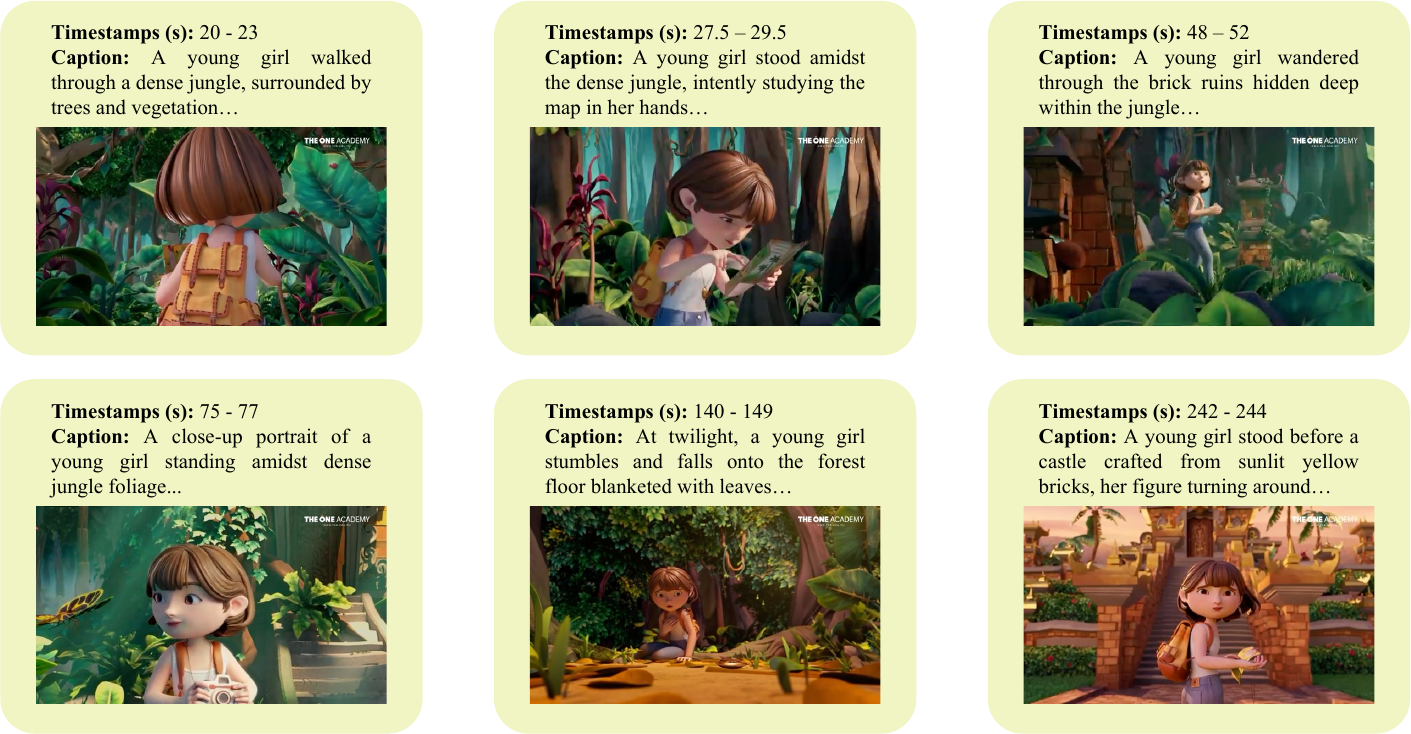}
        \caption{Example of IP-specific dataset for evaluation.}
        \label{fig:app_eval_sample}
        \vspace{-0.5cm}
    \end{center}
\end{figure}

\begin{figure}[htbp]
    \begin{center}
        \includegraphics[width=1\textwidth]{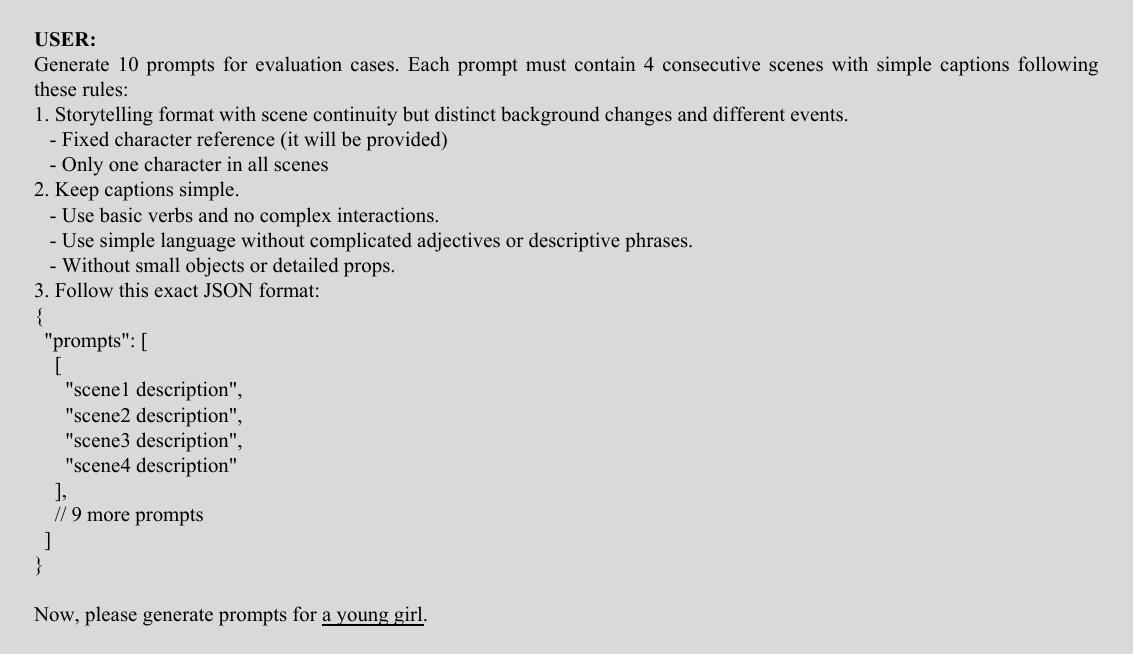}
        \caption{The prompt used for constructing evaluation dataset.}
        \label{fig:app_prompt_eval_sample}
        \vspace{-0.5cm}
    \end{center}
\end{figure}

We construct a manually annotated evaluation dataset of 20 animated films featuring distinct IPs to support LoRA enhancement for AnimeShooterGen and finetuning for CogVideo-LoRA. For each IP, alongside a reference image, we curate 5–6 short video clips (each lasting several seconds) exclusively depicting the main character, ensuring maximal diversity in actions and scenes, as shown in Figure \ref{fig:app_eval_sample}. To evaluate multi-shot generation performance, we employ DeepSeek with prompt shown in Figure \ref{fig:app_prompt_eval_sample} to generate 10 unique narrative prompts per IP.

\subsection{Automatic Metrics}
We employ CLIP score and DreamSim to quantify the visual similarity between generated characters and the reference image. We uniformly sample 5 frames from each shot and compute the average similarity score as the shot-level metric. To assess story-level consistency across 4 shots, we introduce two metrics: (1) Mean Similarity (Mean), the arithmetic mean of the 4 shot-level similarity scores. (2) Penalized Harmonic Mean Similarity (HarMeanP). Recognizing that even a single poorly generated shot can disrupt viewer immersion, this metric penalizes the worst-performing shot. This metric first calculates the harmonic mean of all 4 shots' similarity scores (chosen for its sensitivity to extremely low values), then multiplies this result by the lowest similarity score as an additional penalty term.

\subsection{MLLM Assessment}
\label{sec:app_mllm_eval}
\begin{figure}[h]
    \begin{center}
        \includegraphics[width=1\textwidth]{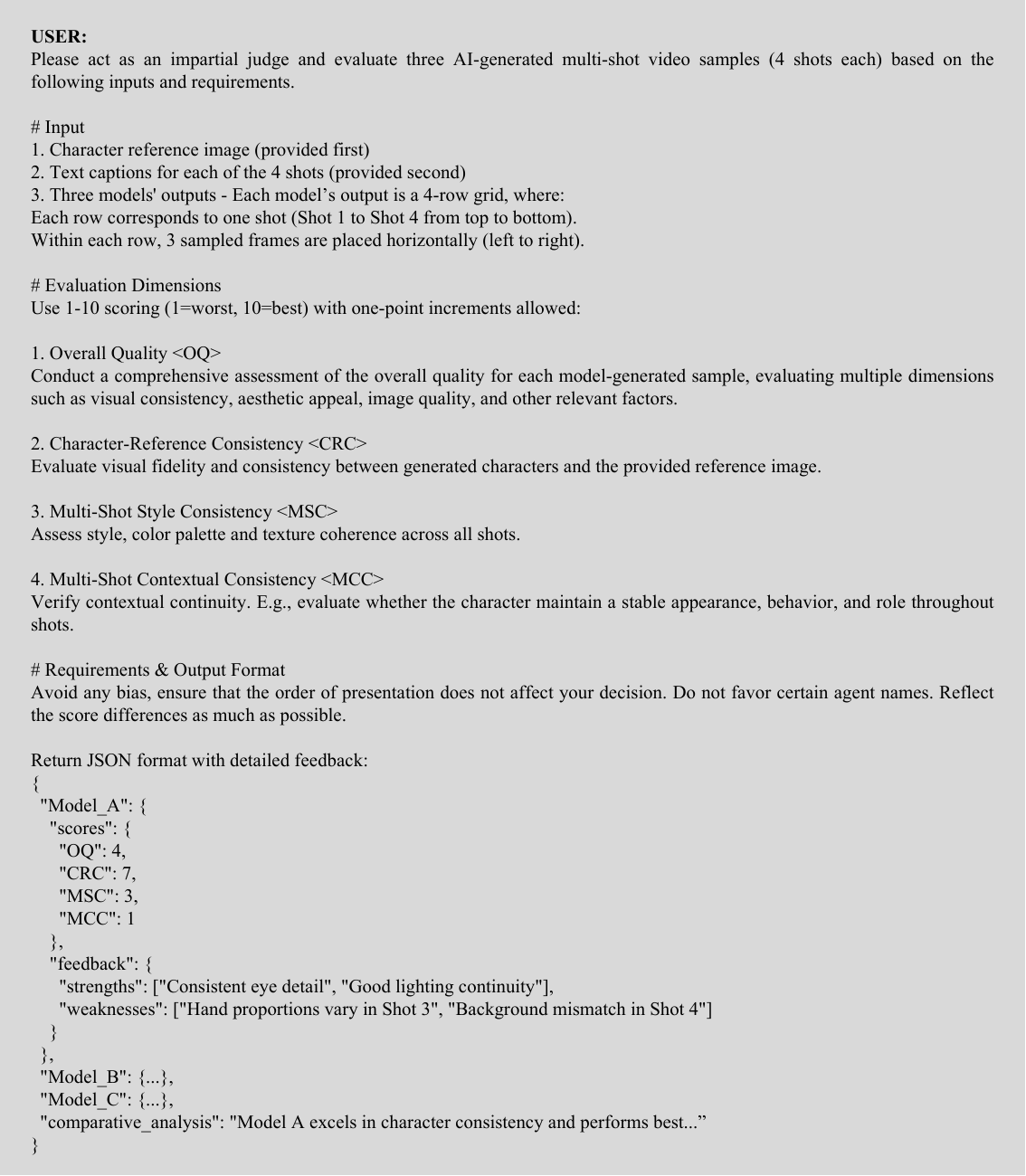}
        \caption{The prompt used for MLLM assessment.}
        \label{fig:app_prompt_mllm_eval}
        \vspace{-0.5cm}
    \end{center}
\end{figure}

We leverage GPT-4o and Gemini 2.5 Pro as MLLM-based judges. To mitigate ordering bias in evaluation, we employ the prompt template shown in Figure \ref{fig:app_prompt_mllm_eval} and change the presentation order across three independent evaluation rounds. The final results reported in Section \ref{sec:quantitative_results} represent the averaged metrics from these three trials, ensuring robustness against positional preferences. MLLM uses 1-10 scoring (1=worst, 10=best) with one-point increments. The evaluation dimensions including:
\begin{itemize}
    \item Overall quality <OQ>. A holistic assessment considering aesthetic appeal, image quality, visual consistency and so on.
    \item Character-Reference Consistency <CRC>. Visual fidelity of the generated character to the provided reference image.
    \item Multi-Shot Style Consistency <MSC>. Coherence of artistic style, color palette, and texture across all shots within a story.
    \item Multi-Shot Contextual Consistency <MCC>. Continuity of the narrative and context across shots, e.g., ensuring the character maintains a consistent appearance appropriate to the unfolding story.
\end{itemize}

\subsection{Human Evaluation}
For the human evaluation, we recruit 10 participants who hold at least a bachelor’s degree and have prior experience in image or video generation. A total of 15 stories with 60 shots are presented to the participants. The evaluation metrics align with those defined in Section \ref{sec:app_mllm_eval}. Participants are instructed to score each dimension (1: lowest, 5: highest) based on the reference image, corresponding captions, and multi-shot videos generated by the three models. The final performance of each model is calculated as the average scores across all responses.

\subsection{Additional Qualitative Results}
\begin{figure}[htbp]
    \begin{center}
        \includegraphics[width=1\textwidth]{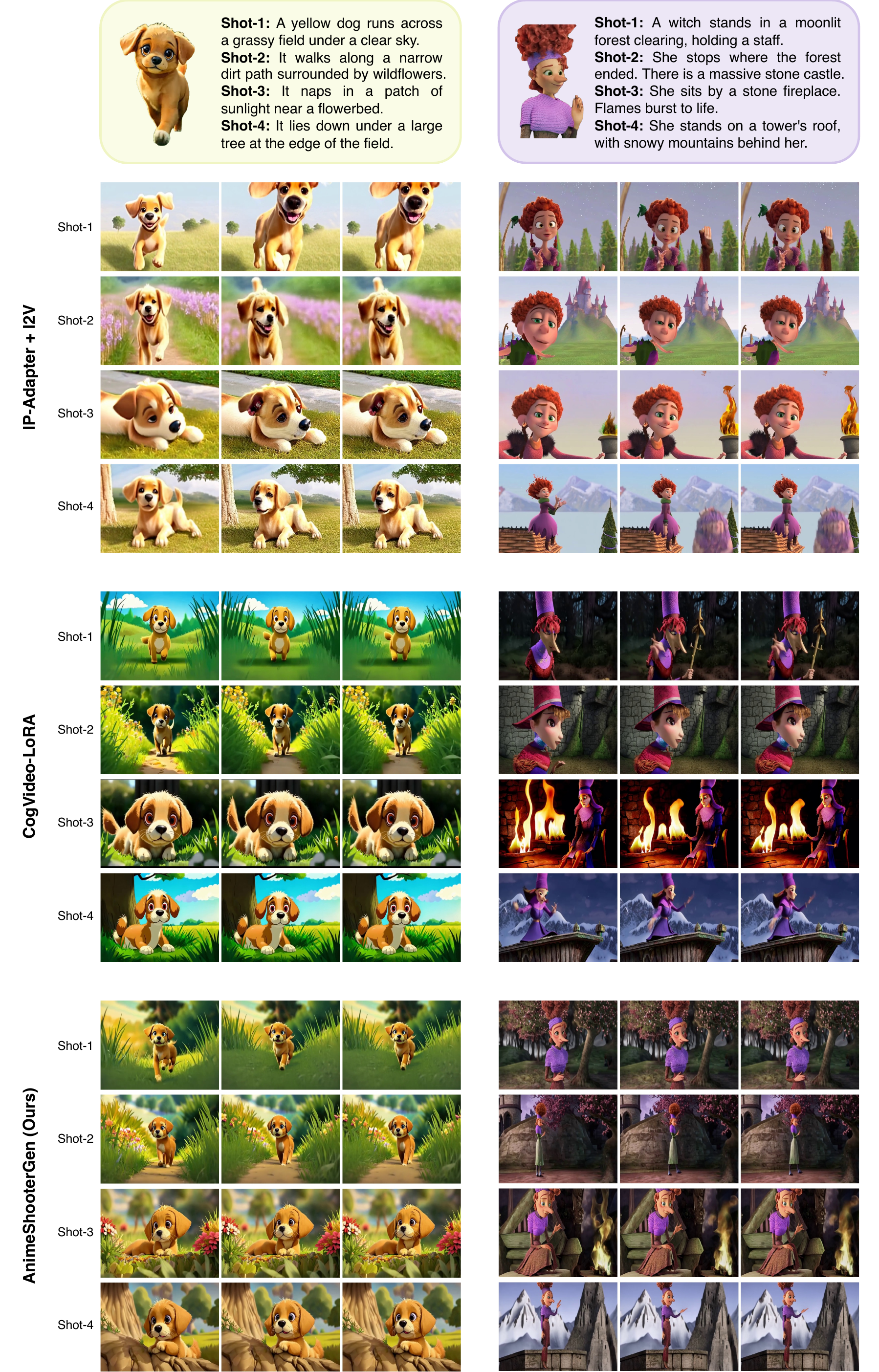}
        \caption{Additional qualitative results.}
        \label{fig:app_results_1}
        \vspace{-0.5cm}
    \end{center}
\end{figure}

\begin{figure}[htbp]
    \begin{center}
        \includegraphics[width=1\textwidth]{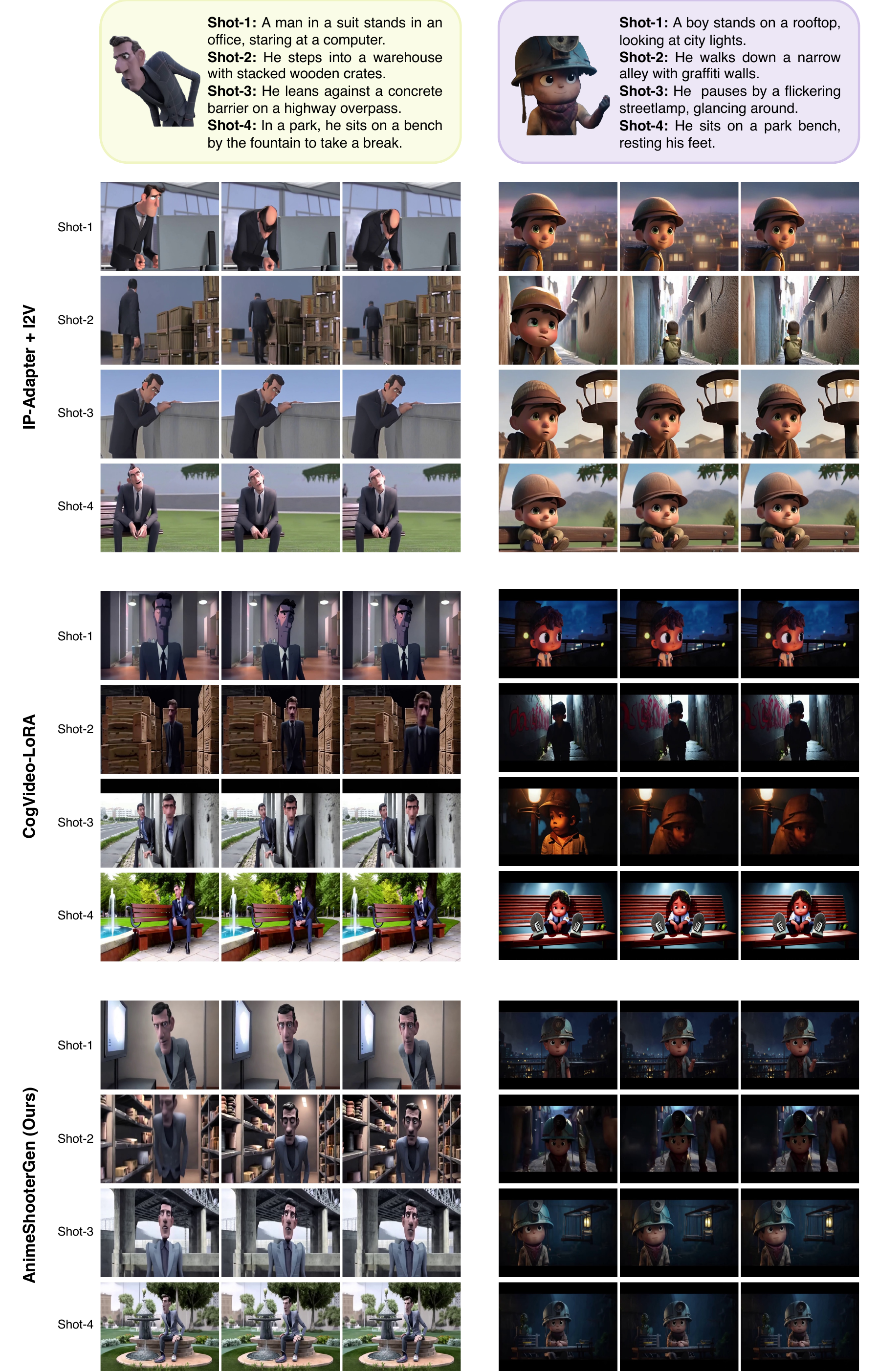}
        \caption{Additional qualitative results.}
        \label{fig:app_results_0}
        \vspace{-0.5cm}
    \end{center}
\end{figure}

We present additional qualitative results in Figure \ref{fig:app_results_1} and Figure \ref{fig:app_results_0}. Both comparison methods exhibit limitations in preserving character consistency with the reference image and cross-shot coherence. As illustrated in the right panel case of Figure \ref{fig:app_results_0}, the IP-Adapter+I2V framework erroneously transforms the flashlight-equipped helmet into a color-mismatched hat. In the left panel case, the first shot generated by CogVideo-LoRA demonstrates a visually discordant art style. In contrast, our proposed AnimeShooterGen achieves superior preservation of character identity, color palette continuity, and stylistic consistency across generated sequences.

\section{Integration of Audio Generation Capabilities for AnimeShooterGen}
\begin{figure}[htbp]
    \begin{center}
        \includegraphics[width=1\textwidth]{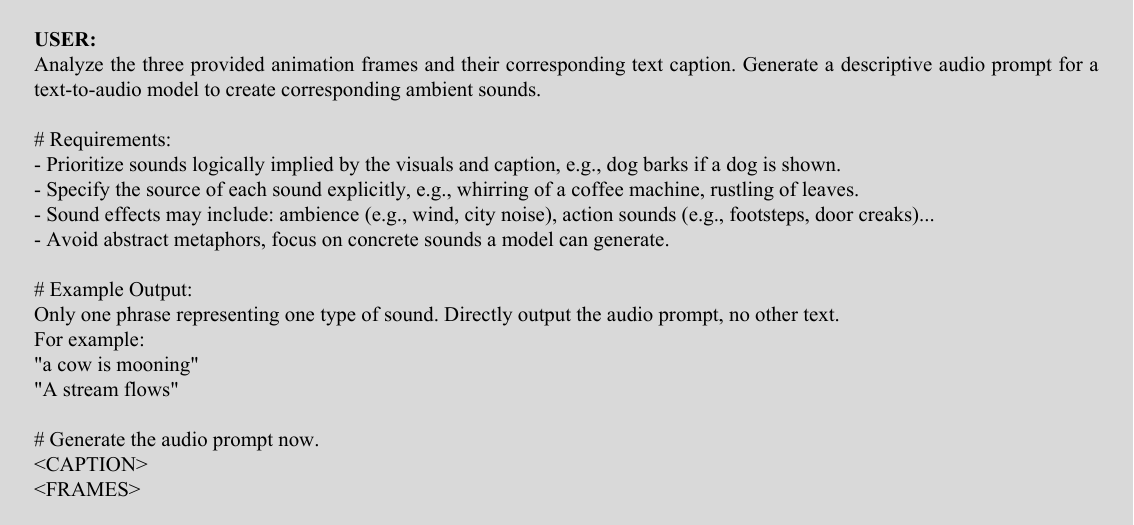}
        \caption{The prompt used for generating descriptive audio captions.}
        \label{fig:app_prompt_tangoflux}
        \vspace{-0.5cm}
    \end{center}
\end{figure}

To further augment the immersive quality, we integrate AnimeShooterGen with zero-shot Text-to-Audio (TTA) generation using TangoFlux \cite{hung2024tangoflux}. The workflow involves processing the video captions and keyframes with GPT-4o to generate descriptive audio captions with the prompt shown in Figure \ref{fig:app_prompt_tangoflux}. These audio captions subsequently guide TangoFlux in synthesizing audio tracks, which are then merged with the video sequences. However, results reveal substantial limitations in current simplistic zero-shot audio generation paradigms. Primarily, the decoupled generation processes for visual and auditory modalities result in inherent inter-modality synchronization failures. For example, footstep sounds lag behind walking animations, or character facial expressions mismatch with voice. Furthermore, constrained by existing text-to-audio models' performance, environmental sound effects such as gentle breezes, engine roars, and mechanical hums fail to achieve sufficient perceptual distinctiveness, thereby compromising the immersive experience. We propose an audio-annotated subset named AnimeShooter-audio, hoping to facilitate and encourage future research into the development of more sophisticated audio-visual co-generation models capable of achieving tighter synchronization and semantic coherence.

\section{Potential Negative Social Impacts}
While animation generation models democratize content creation, they risk enabling malicious applications such as generating deepfakes for disinformation, infringing intellectual property, or producing harmful content. To address these risks, the research community and policymakers must adopt proactive safeguards. Technical countermeasures include embedding watermarks in generated videos for provenance tracking, deploying AI-driven detectors to flag synthetic content, and implementing strict input/output filters to block unethical prompts.

\section{Data Access and License}
The AnimeShooter and AnimeShooter-audio datasets are released under the CC BY-NC 4.0 License and the data is collected from publicly accessible sources. These released datasets include annotated scripts, reference image masks, and corresponding video IDs, while the original source videos must be obtained independently from YouTube using the provided IDs. AnimeShooterGen is built upon two pretrained models: NVILA~\cite{liu2024nvila} and CogVideo~\cite{hong2022cogvideo}. We will release the code and model weights of AnimeShooterGen under the Apache 2.0 License, and provide a copy of the original licenses of NVILA and CogVideo in our GitHub repository.

To utilize the AnimeShooter and AnimeShooter-audio datasets, users must first download the source videos from YouTube via the specified video IDs or URLs using the yt-dlp tool. Subsequently, video segments should be extracted based on the provided start and end frame indices through ffmpeg processing. The extracted segments can then be analyzed using the accompanying story-level and shot-level annotations. Additionally, binary masks and corresponding frame indices are supplied for reference image generation. These reference images can be obtained by applying the provided masks to the cropped video frames through pixel-wise multiplication.

\end{document}